\documentclass[pmlr]{jmlr}
\usepackage{wrapfig}
\usepackage{booktabs}
\usepackage{graphicx}
\newcommand{\tagdsmode}{proceedings}

\makeatletter
\newcommand{\tagdssubmission}{submission}
\newcommand{\tagdsproceedings}{proceedings}

\ifx\tagdsmode\tagdsproceedings

\else\ifx\tagdsmode\tagdssubmission
  \def\ps@jmlrtps{%
    \let\@mkboth\@gobbletwo
    \def\@oddhead{\scriptsize Under Review at the 2nd Conference on Topology, Algebra, and Geometry in Data Science\hfill}%
    \let\@evenhead\@oddhead
    \def\@oddfoot{}%
    \let\@evenfoot\@oddfoot
  }

\else
  \def\ps@jmlrtps{%
    \let\@mkboth\@gobbletwo
    \def\@oddhead{}%
    \let\@evenhead\@oddhead
    \def\@oddfoot{}%
    \let\@evenfoot\@oddfoot
  }
\fi\fi
\makeatother

\usepackage{longtable}%

\usepackage{booktabs}
\usepackage[load-configurations=version-1]{siunitx} %
\usepackage{bbm}            %
\usepackage{adjustbox}
\usepackage{algpseudocode}

\usepackage{amsmath, amssymb}
\usepackage{mathtools}  %
\usepackage{bm}         %

\providecommand{\R}{\mathbb{R}}
\providecommand{\V}{\mathcal{V}}
\providecommand{\X}{\mathcal{X}}

\providecommand{\x}{\mathbf{x}}

\providecommand{\z}{\mathbf{z}}

\providecommand{\I}{\mathbbm{1}}
\renewcommand{\u}{\mathbf{u}}   %
\newcommand{\h}{\mathbf{h}}
\providecommand{\FDD}{\textsc{FDD}}
\providecommand{\Enc}{\mathrm{Enc}}
\providecommand{\Dec}{\mathrm{Dec}}

\usepackage{cleveref}       %
\crefname{algocf}{Algorithm}{Algorithms}   %
\Crefname{algocf}{Algorithm}{Algorithms}

\theorembodyfont{\upshape}
\theoremheaderfont{\scshape}
\theorempostheader{:}
\theoremsep{\newline}

\jmlrvolume{334}
\jmlryear{2026}
\jmlrworkshop{Topology, Algebra, and Geometry in Data Science}

\title[Freeze, Diffuse, Decode]{Freeze, Diffuse, Decode: Task-Aware Adaptation of Transformer Embeddings for Antimicrobial Peptide Design}

\ifx\tagdsmode\tagdssubmission

\else
  \author{\Name{Pankhil Gawade\textsuperscript{*}} \Email{gawade.pankhil@helmholtz-munich.de} \\
        \addr Institute of AI for Health, Helmholtz Zentrum Munchen\\
        \addr Technical University of Munich, TUM School of Computation, Information and Technology\\
        \\
   \Name{Adam Izdebski\textsuperscript{*}} \Email{}\\
        \addr Institute of AI for Health, Helmholtz Zentrum Munchen\\
        \addr Technical University of Munich, TUM School of Computation, Information and Technology\\
        \\
   \Name{Myriam Lizotte} \Email{}\\
   \addr Mila – Quebec AI Institute, Montreal, Canada \\
   \addr Department of Mathematics and Statistics, Université de Montreal\\
   \\
   \Name{Kevin R. Moon} \Email{}\\
   \addr Dept. of Mathematics \& Statistics Utah State University, Logan, UT, USA\\
   \\
   \Name{Jake S. Rhodes} \Email{}\\
   \addr Dept. of Statistics Brigham Young University, Provo, UT, USA\\
   \\
   \Name{Guy Wolf}\Email{}\\
   \addr Mila – Quebec AI Institute, Montreal, Canada \\
   \addr Department of Mathematics and Statistics, Université de Montreal\\ 
   \\
   \Name{Ewa Szczurek} \Email{ewa.szczurek@helmholtz-munich.de}\\
   \addr Institute of AI for Health, Helmholtz Zentrum Munchen\\
   \addr Faculty of Mathematics, Informatics and Mechanics, University of Warsaw }

\fi

\begin{document}

\maketitle
\begingroup
\renewcommand{\thefootnote}{*}
\footnotetext{Equal contribution.}
\endgroup
\begin{abstract}
Pretrained transformers provide general-purpose molecular embeddings for downstream tasks. While these embeddings provide task-agnostic structural patterns, they lack task-specific alignment, %
limiting downstream performance. %
Here, we introduce \textsc{Freeze, Diffuse, Decode} ($\FDD$), a diffusion-based framework that adapts pretrained transformer embeddings to downstream tasks by building a task-supervised diffusion geometry over the frozen embeddings, without any backbone training. 
Applied to antimicrobial peptide design, \FDD{} yields low-dimensional, predictive, and interpretable representations that support property prediction, retrieval, and latent-space~interpolation.
\end{abstract}
\begin{keywords}
molecular representation learning, manifold learning, diffusion maps, antimicrobial peptide design
\end{keywords}

\section{Introduction}

Molecular representation learning is fundamental to modern drug discovery. Embedding molecules into high-dimensional features that capture general structural properties enables downstream tasks such as molecular property prediction, molecule retrieval, and chemical space exploration. Recently, pretrained transformers have been reshaping molecular machine learning by providing embeddings that are increasingly leveraged across downstream tasks~\citep{chithrananda2020,zhou2023unimol,lin2023esm2}.

However, pretrained embeddings underperform on downstream tasks. In particular, probing pretrained embeddings for molecular property prediction is often outperformed by hand-crafted descriptors~\citep{praski2025benchmarkingpretrainedmolecularembedding,adamczyk2025molecularfingerprintsstrongmodels}. A possible reason is that pretrained embeddings are organized around general structural similarity, lacking task-specific alignment. While fine-tuning offers strong task-specific alignment, it inflates the representation's intrinsic dimensionality, a correlate of poorer generalization~\citep{ansuini2019intrinsicdimensiondatarepresentations,valeriani2023geometryhiddenrepresentationslarge}, and is sample-inefficient, offering a poor fit for low-data regimes typical of antimicrobial peptide design~\citep{VANTILBORG2024102818}.
To address this gap, we introduce \textsc{Freeze, Diffuse, Decode} (\FDD), a framework based on diffusion geometry~\citep{coifman2006diffusion} that adapts frozen transformer embeddings to a downstream task without fine-tuning. \FDD{} constructs a task-supervised diffusion geometry over the embeddings rather than raw inputs~\citep{rhodes2023rfphate,aumon2025rfae}, then trains an encoder to extend it to unseen molecules. The resulting representation is low-dimensional and task-aware, improving over the frozen embeddings and narrowing the gap to fine-tuning. Our contributions are:
\begin{itemize}
\item We introduce \FDD, which adapts frozen transformer embeddings to a downstream task through supervised diffusion geometry, without retraining the backbone.
\item We show that \FDD{} yields low-dimensional, predictive, and interpretable representations for molecular property prediction and peptide retrieval.
\item We show that \FDD{} supports latent-space interpolation whose decoded peptides gradually acquire physicochemical traits of antimicrobial activity.
\end{itemize}

\section{Freeze, Diffuse, Decode}
\label{sec:methods}

We propose \textsc{Freeze, Diffuse, Decode} ($\FDD$), a framework for adapting frozen embeddings of a pretrained molecular transformer to downstream tasks via supervised diffusion geometry. Our framework runs in three stages. We first \emph{freeze} a pretrained transformer and extract its embeddings (\Cref{sec:freeze}). We then \emph{diffuse} by building a supervised diffusion geometry over the frozen embeddings and training an encoder to map any molecule into low-dimensional geometry-aware representations (\Cref{sec:diffuse}). We finally \emph{decode} these representations into task-specific outputs, fitting a lightweight predictor for property prediction and retrieval or decoding back to sequence space for generation (\Cref{sec:decode}).

\subsection{Freeze}
\label{sec:freeze}
 
The \emph{Freeze} stage extracts embeddings from a pretrained transformer.
Let $\V$ be a vocabulary of tokens, for example amino acids or the characters of a SMILES string. A molecule $\x = (x_1,\dots,x_T) \in \V^T$ is a finite sequence of such tokens, and the design space $\X = \V^*$ collects all finite-length sequences over $\V$. A frozen, pretrained transformer
\begin{equation}
    f \colon \X \to \R^D
\end{equation}
maps each molecule to a $D$-dimensional embedding $\h = f(\x)$, where $D$ is typically large.

\subsection{Diffuse}
\label{sec:diffuse}

The \emph{Diffuse} stage adapts the frozen embeddings to a downstream task by learning a mapping
\begin{equation}
    g_\theta \colon \R^D \to \R^d, \qquad d \ll D,
\end{equation}
so that $\z \coloneqq (g_\theta \circ f)(\x) \in \R^d$ is low-dimensional, yet informative for the downstream task. We construct a label-aware diffusion geometry on the training embeddings, then fit a parametric autoencoder that extends that geometry to molecules unseen during training.

\paragraph{Supervised diffusion geometry}
We measure similarity between frozen embeddings with RF-GAP proximities~\citep{rhodes2023rfgap}, which treat two points as close when a random forest trained to predict $y$ routes them to the same leaves. Specifically, we fit a random forest to the labeled pairs $\mathcal{D} = \{(\h_n, y_n)\}_{n \in [N]}$, where $\h_n = f(\x_n)$ is the embedding of molecule $\x_n$ and $[N] = \{1,\dots,N\}$. Each tree $t$ is grown on a bootstrap sample $B(t) \subseteq \mathcal{D}$ of size $N$, drawn with replacement. We write $c_j(t)$ for the number of times $\h_j$ is drawn into $B(t)$, $S_i = \{t : \h_i \notin B(t)\}$ and $\bar{S}_i = \{t : \h_i \in B(t)\}$ for the sets of trees in which $\h_i$ is out-of-bag and in-bag, respectively, and $J_i(t)$ for the in-bag points that fall in the same leaf as $\h_i$ in tree $t$. The leaf holding $\h_i$ in tree $t$ contains $|M_i(t)| = \sum_j c_j(t)\,\I\!\left(\h_j \in J_i(t)\right)$ in-bag points in total, counting each $\h_j$ as many times as it is drawn into $B(t)$. For every ordered pair $(i,j) \in [N]\times[N]$, we define RF-GAP proximity as
\begin{equation}
    p_{\mathrm{GAP}}(i,j) =
    \begin{cases}
        \dfrac{1}{|\bar{S}_i|}\displaystyle\sum_{t \in \bar{S}_i} \dfrac{c_i(t)}{|M_i(t)|}, & i = j,\\[2.4ex]
        \dfrac{1}{|S_i|}\displaystyle\sum_{t \in S_i} \dfrac{c_j(t)\,\I\!\left(\h_j \in J_i(t)\right)}{|M_i(t)|}, & i \neq j.
    \end{cases}
    \label{eq:rfgap}
\end{equation}
Restricting the off-diagonal sum to out-of-bag trees makes RF-GAP proximities geometry- and accuracy-preserving, in the sense that a proximity-weighted nearest-neighbor predictor reproduces the random forest's out-of-bag predictions~\citep{rhodes2023rfgap}.

Next, we collect pairwise embedding proximities into a matrix $\mathbf{W} \in \R^{N\times N}$ with entries $\mathbf{W}_{ij} = p_{\mathrm{GAP}}(i,j)$ for $i, j \in [N]$. We make it symmetric, $\tfrac12(\mathbf{W}+\mathbf{W}^\top)$, and row-normalize the result into a row-stochastic diffusion operator $\mathbf{P} \in \R^{N\times N}$ that defines a Markov chain over $\mathcal{D}$. Thus, each entry $\mathbf{P}_{ij}$ is the one-step transition probability from $\h_i$ to $\h_j$, and each row $\mathbf{P}_i$ captures the local geometry around $\h_i$.

Finally, following RF-PHATE~\citep{rhodes2023rfphate}, we diffuse the operator $\mathbf{P}$ by powering it to a diffusion time $t$, which accumulates every length-$t$ transition path and so integrates local proximities into global structure~\citep{coifman2006diffusion}. We choose $t$ large enough to smooth away local noise yet small enough to preserve the manifold's structure, deferring the selection criterion to~\Cref{apd:method-details}. We apply the potential transform $U_t = -\log \mathbf{P}^t$ elementwise, defining the \emph{potential distance} $V_t(i,j) = \lVert \u_i - \u_j\rVert_2$ between its rows $\u_i$ and $\u_j$ and embed these distances into $\R^d$ by metric multidimensional scaling, yielding supervised target coordinates $\z^{G}_i \in \R^d$~\citep{Moon2019-ue}. 

\paragraph{Landmarks}
The potential-distance MDS over all $N$ training points dominates the cost of building the supervised targets, scaling as $O(N^3)$ in time and $O(N^2)$ in memory. To scale to large $N$, we follow PHATE~\citep{Moon2019-ue} and compute the diffusion geometry on $M \ll N$ landmarks, embedding the landmarks with MDS and extending the coordinates to the remaining points through their transition probabilities to the landmarks. This lowers the dominant cost to $O(M^3 + NM)$ time and $O(NM)$ memory.

\paragraph{Geometry-regularized autoencoder}
To extend supervised target coordinates $\z^{G}_i$ out-of-sample, i.e., beyond the training dataset, we follow RF-AE~\citep{aumon2025rfae} and train an autoencoder that reconstructs transition vectors $p_i \in \Delta^{N-1}$ encoding the supervised geometry. The encoder $\Enc_\theta(p_i) = \z_i \in \R^d$ compresses a transition vector $p_i \in \Delta^{N-1}$, and the decoder $\Dec_\phi(\z_i) = \hat{p}_i \in \Delta^{N-1}$ returns a transition distribution through a softmax output layer, while a geometric term anchors the bottleneck to the precomputed RF-PHATE coordinates~\citep{duque2023grae}. We fit $(\theta,\phi)$ to minimize
\begin{equation}
    \mathcal{L}(\theta,\phi) = \frac{1}{N}\sum_{i=1}^{N} \Bigl[\, (1-\lambda)\, D_{\mathrm{KL}}(p_i \,\|\, \hat{p}_i) + \lambda\, \lVert\z_i - \z^{G}_i\rVert_2^2 \,\Bigr],
    \label{eq:rfae-loss}
\end{equation}
where the reconstruction term is the Kullback--Leibler divergence between the transition distributions $p_i$ and $\hat{p}_i$, and the geometric term regresses $\z_i$ onto its RF-PHATE target $\z^G_i$. The weight $\lambda \in [0,1]$ interpolates between a plain autoencoder ($\lambda = 0$) and pure regression onto the RF-PHATE embedding ($\lambda = 1$).

\subsection{Decode}
\label{sec:decode}
 
The \emph{Decode} stage maps latent representations $\z$ to a downstream task-specific output.
 
\paragraph{Prediction and retrieval}
For prediction and retrieval, we probe $\z$ by $k$-NN or fitting a linear prediction head on the latent representations, mapping a molecule to its target $\hat{y}$.
 
\paragraph{Generation}
The autoencoder decoder $\Dec_\phi$ maps a latent representation to a reconstructed proximity vector $\hat{p} = \Dec_\phi(\z) \in \R^N$; a learned map $m_\xi \colon \R^N \to \R^D$ then lifts these proximities to the frozen embedding space, $\hat{\h} = m_\xi(\hat{p})$, and the frozen transformer's own decoder maps $\hat{\h}$ back to a sequence in $\X$. %

\paragraph{Latent space interpolation via neural ODEs}

To explore continuous interpolation in the latent space, we define an ordinary differential equation (ODE) over the diffusion coordinates $\boldsymbol{z}(t)\in\mathbb{R}^d$
\[
\frac{d\boldsymbol{z}(t)}{dt} = v_\psi(\boldsymbol{z}(t),t),
\]
where $v_\psi \colon \mathbb{R}^d \times[0,1] \rightarrow \mathbb{R}^d $ is a learnable velocity field parameterized by a neural network with weights $\psi$, trained with Sinkhorn loss~\citep{cuturi2013sinkhorndistanceslightspeedcomputation,sinkhorn2018pmlr}. Solving the above ODE produces smooth interpolation paths $\{\boldsymbol{z}(t) \mid t \in [0, 1]\}$ in the latent space, enabling transformation of latent representations from one region to the other. Specifically, at each training iteration, we sample a mini-batch \(\{\boldsymbol{z}_m(0)\}_{m=1}^{M}\) from one region in the latent space and a target mini-batch \(\{\boldsymbol{z}_{m}(1)\}_{m=1}^{M}\) from the other. Next, we compute trajectory points \(\boldsymbol{z}_m(t)\) via the ODE solver and evaluate the accelerated Sinkhorn Loss \(\mathcal{L}_{\mathrm{Sinkhorn}}\), with \(\varepsilon=0.1\) regularization for stable gradients, given by 
\begin{equation}
\mathcal{L}_{\mathrm{Sinkhorn}}
=
\mathrm{OT}_\varepsilon(\mu,\nu)
\;-\;\tfrac{1}{2}\Bigl[\mathrm{OT}_\varepsilon(\mu,\mu)
+\mathrm{OT}_\varepsilon(\nu,\nu)\Bigr],
\end{equation}
where $\mathrm{OT}_\varepsilon$ is the entropy-regularized optimal transport term \begin{equation}
\mathrm{OT}_\varepsilon(\mu,\nu)
=
\min_{P\in\Pi(\mu,\nu)}
\left[
\sum_{i,j} P_{ij}\,\|\boldsymbol{z}_i - \boldsymbol{z}_j\|^2 + \varepsilon \,  \text{KL}(P \| \mu \otimes \nu)
\right] ,
\end{equation} where 
\( \mu, \nu\) are respectively the source and target marginal probability distributions
(in our case, uniform distributions over the sample trajectory points \(\{\boldsymbol{z}_m(0)\}\) or the targets \(\{\boldsymbol{z}_m(1)\}\) respectively),
\(\{\boldsymbol{z}_i\}\sim\mu\) and \(\{\boldsymbol{z}_j\}\sim\nu\) are samples from the two probability distributions, 
$P$ is a matrix representing a transport plan subject to marginal constraints defined by $\Pi$,
and \(\text{KL}(P \| \mu \otimes \nu)\) denotes the KL-divergence of $P$ with respect to the product measure $\mu \otimes \nu$. Finally, we jointly optimize the neural network parameters by backpropagating through both the ODE integration (adjoint sensitivity) and the Sinkhorn computation. Once trained, latent codes $\boldsymbol{z}_m(t)$ are passed through our decoding pipeline, yielding novel peptide sequences.  
 
\subsection{Inference}
At inference, we route a molecule's frozen embedding $\h = f(\x)$ through the fitted random forest, applying~\Cref{eq:rfgap} with every tree being out-of-bag, to obtain a transition vector $\mathbf{p} \in \Delta^{N-1}$ that the encoder maps to a latent representation $\z = \Enc_\theta(\mathbf{p})$. The task-specific head then decodes $\z$ to a desired output. %
 
\section{Experiments}

To demonstrate the utility of adapting pretrained transformer embeddings to downstream tasks in antimicrobial peptide (AMP) discovery, we benchmark \FDD{} on property prediction (Section~\ref{section:experimental-results:prediction} and~\ref{section:experimental-results:id}), peptide retrieval (Section~\ref{section:experimental-results:retrieval}), and latent-space interpolation (Section~\ref{section:experimental-results:interpolation}). Our results indicate that \FDD{} yields task-aware representations that improve over the pretrained embeddings across all three settings. Additionally, we report all experimental details, ablations and additional results in Appendix~\ref{apd:method-details}, \ref{sec:ablations} and~\ref{apd:first}, respectively.

\subsection{\FDD{} outperforms pretrained embeddings on property prediction}\label{section:experimental-results:prediction}

To evaluate whether the task-aware representations produced by \FDD{} improve downstream
prediction, we consider binary antimicrobial activity classification across \textit{ESKAPE}
bacterial species from~\citet{10.1093/nar/gkaa991} (see Appendix~\ref{data}). We compare
embeddings extracted from two pretrained backbones: Hyformer~\citep{izdebski2025synergisticbenefitsjointmolecule}
($33$M parameters) and ESM-2~\citep{lin2023esm2} ($35$M parameters). For each backbone we probe the pretrained and \FDD{} embeddings
with a linear classifier; end-to-end fine-tuning of the backbone serves as an upper-bound
reference, and a vanilla autoencoder (AE) as an unsupervised dimensionality-reduction
control.

\begin{table*}[ht!]
\caption{Predictive performance on antimicrobial activity classification across ESKAPE
bacterial species. Among non-fine-tuned methods, the best embedding per species is marked
\textbf{bold}; fine-tuning is reported separately as an upper-bound reference. Mean (std)
across 10 seeds; pretrained is deterministic.\label{tab:hyformer_esm2_auc}}
\vspace{-2ex}
\begin{center}
\begin{scriptsize}
\begin{sc}
\begin{adjustbox}{width=\linewidth}
\begin{tabular}{@{}lcccccc@{}}
\toprule
& \multicolumn{6}{c}{Bacterial Species, AUC ($\uparrow$)} \\
\cmidrule(lr){2-7}
Method & \textit{A. baumannii} & \textit{E. coli} & \textit{E. faecium} & \textit{K. pneumoniae} & \textit{P. aeruginosa} & \textit{S. aureus} \\
\midrule

\textbf{Hyformer} \\
\quad Pretrained        & 0.843 & 0.882 & 0.800 & 0.826 & 0.842 & 0.823 \\
\quad AE                & 0.780 (0.006) & 0.826 (0.004) & 0.752 (0.016) & 0.774 (0.009) & 0.747 (0.003) & 0.764 (0.005) \\
\quad \FDD{} (ours)     & \textbf{0.871 (0.013)} & \textbf{0.864 (0.005)} & \textbf{0.834 (0.014)} & \textbf{0.828 (0.010)} & \textbf{0.859 (0.004)} & \textbf{0.832 (0.005)} \\
\cmidrule(lr){2-7}
\quad Fine-tuning       & 0.859 (0.005) & 0.885 (0.005) & 0.821 (0.019) & 0.841 (0.014) & 0.854 (0.010) & 0.846 (0.004) \\
\midrule
\textbf{ESM-2} \\
\quad Pretrained        & 0.828 & 0.865 & 0.750 & 0.799 & 0.818 & 0.809 \\
\quad AE                & 0.802 (0.012) & 0.816 (0.003) & 0.803 (0.020) & 0.753 (0.002) & 0.744 (0.004) & 0.766 (0.001) \\
\quad \FDD{} (ours)     & \textbf{0.852 (0.008)} & \textbf{0.866 (0.006)} & \textbf{0.825 (0.016)} & \textbf{0.829 (0.012)} & \textbf{0.856 (0.003)} & \textbf{0.837 (0.007)} \\
\cmidrule(lr){2-7}
\quad Fine-tuning       & 0.840 (0.003) & 0.898 (0.006) & 0.832 (0.009) & 0.853 (0.002) & 0.856 (0.010) & 0.854 (0.005) \\
\bottomrule
\end{tabular}
\end{adjustbox}
\end{sc}
\end{scriptsize}
\end{center}
\vspace{-2ex}
\end{table*}

\textsc{FDD} improves the predictive performance of raw pretrained embeddings for both Hyformer and ESM-2 (Table~\ref{tab:hyformer_esm2_auc}), raising the average AUC from $0.836$ to
$0.848$ for Hyformer and from $0.812$ to $0.844$ for ESM-2, with a per-species gain in
every case except for Hyformer on \textit{E. coli}. In contrast, unsupervised autoencoder (AE) degrades performance across all species and backbones (down to $0.774$ and $0.781$ on average), showing that embedding compression without supervised diffusion geometry may discard useful structure. Fine-tuning attains the highest average AUC overall
($0.851$ for Hyformer, $0.856$ for ESM-2), but \textsc{FDD} recovers most of this gap
without updating the backbone.
On a per-species basis \textsc{FDD} matches or exceeds fine-tuning on three of six species
for Hyformer (\textit{A. baumannii}, \textit{E. faecium}, \textit{P. aeruginosa}) and on
(\textit{A. baumannii} and \textit{P. aeruginosa})  for ESM-2. Overall, \textsc{FDD} extracts task-relevant structure from frozen pretrained
embeddings, recovering most of the fine-tuning expressivity.

\subsection{\FDD{} adapts without inflating intrinsic dimensionality}\label{section:experimental-results:id}

Intrinsic dimensionality (ID) measures the effective number of dimensions a representation uses, the dimension of the manifold on which embeddings concentrate and typically well below the ambient width $D$. It is a useful proxy for adaptation cost. Lower terminal-layer ID correlates with better generalization~\citep{ansuini2019intrinsicdimensiondatarepresentations,valeriani2023geometryhiddenrepresentationslarge}, while fine-tuning typically realigns the backbone by recruiting additional directions, inflating ID, precisely the kind of added complexity that drives overfitting in the low-data AMP regime. An ideal adaptation improves task alignment while keeping the manifold compact.

We estimate local ID via the participation ratio of the local PCA spectrum over active peptides ($\mathrm{MIC}<32\,\mu$M), in three per-species spaces: frozen pretrained Hyformer (PT), the \FDD{} projection of PT, and fine-tuned Hyformer (FT). Full estimator and subset details are in Appendix~\ref{apd:id}.

\begin{table*}[ht!]
\caption{Estimated local intrinsic dimensionality (ID) across ESKAPE bacterial species. Avg.\ is the cross-species mean.}\label{tab:ID}
\vspace{-2ex}
\begin{center}
\begin{scriptsize}
\begin{sc}
\begin{adjustbox}{width=\linewidth}
\begin{tabular}{@{}lccccccc@{}}
\toprule
& \multicolumn{6}{c}{Bacterial Species, ID ($\downarrow$)} & \\
\cmidrule(lr){2-7}
Method & \textit{A. baumannii} & \textit{E. coli} & \textit{E. faecium} & \textit{K. pneumoniae} & \textit{P. aeruginosa} & \textit{S. aureus} & Avg. \\
\midrule
Pretrained     & $6.10$ & $6.14$ & $6.25$ & $6.05$ & $6.18$ & $6.33$ & $6.18$ \\
\FDD{} (ours)   & $5.62$ & $6.44$ & $4.83$ & $5.72$ & $6.67$ & $6.46$ & $5.96$ \\
Fine-tuning     & $7.43$ & $8.24$ & $6.57$ & $7.04$ & $7.84$ & $8.28$ & $7.57$ \\
\bottomrule
\end{tabular}
\end{adjustbox}
\end{sc}
\end{scriptsize}
\end{center}
\vspace{-2ex}
\end{table*}

Fine-tuning raises the average ID from $6.18$ (Pretrained) to $7.57$, inflating it across all species. \FDD{} instead holds ID near the pretrained manifold (average $5.96$) and below FT in all six cases. \FDD{} thus improves downstream predictive performance without the dimensional inflation of fine-tuning, consistent with a compact, task-aware reorganization of the frozen geometry rather than a wholesale expansion of it.

\subsection{\FDD{} retrieves novel active peptides}\label{section:experimental-results:retrieval}

\begin{wraptable}{r}{0.3\linewidth}
\vspace{-20pt}
\centering
\caption{AMP retrieval performance.\label{tab:test_metrics}}
\small
\setlength{\tabcolsep}{3pt}
\renewcommand{\arraystretch}{0.95}
\begin{tabular}{@{}lc@{}}
\toprule
\textbf{Method} & \textbf{$P_{\mathrm{amp}}$ ($\uparrow$)} \\
\midrule
RF          & 0.80 \\
HydrAMP     & 0.83 \\
AMPScanner  & 0.89 \\
AMPlify     & 0.89 \\
\textbf{FDD (ours)} & \textbf{0.90} \\
\bottomrule
\end{tabular}
\vspace{-10pt}
\end{wraptable}

We test whether \FDD{} embeddings capture biological similarity well enough to retrieve novel actives. We fit a $k$NN classifier ($k{=}5$) on \FDD{} embeddings of the~\citet{Veltri2018-pl} dataset and use it to rank peptides in the held-out classification set of~\citet{Szymczak2023-hf}, none seen during training. We compare against state-of-the-art AMP classifiers, a random forest on physicochemical descriptors (RF), HydrAMP~\citep{Szymczak2023-hf}, AMPScanner~\citep{YANG}, and AMPlify~\citep{Li2022-gq} on the same held-out set, reporting retrieval precision $P_{\mathrm{amp}}$ ($\uparrow$).

\begin{figure}[htbp]
\centering
\vspace{-10pt}
\begin{minipage}[t]{0.48\linewidth}
    \centering
    \includegraphics[width=\linewidth]{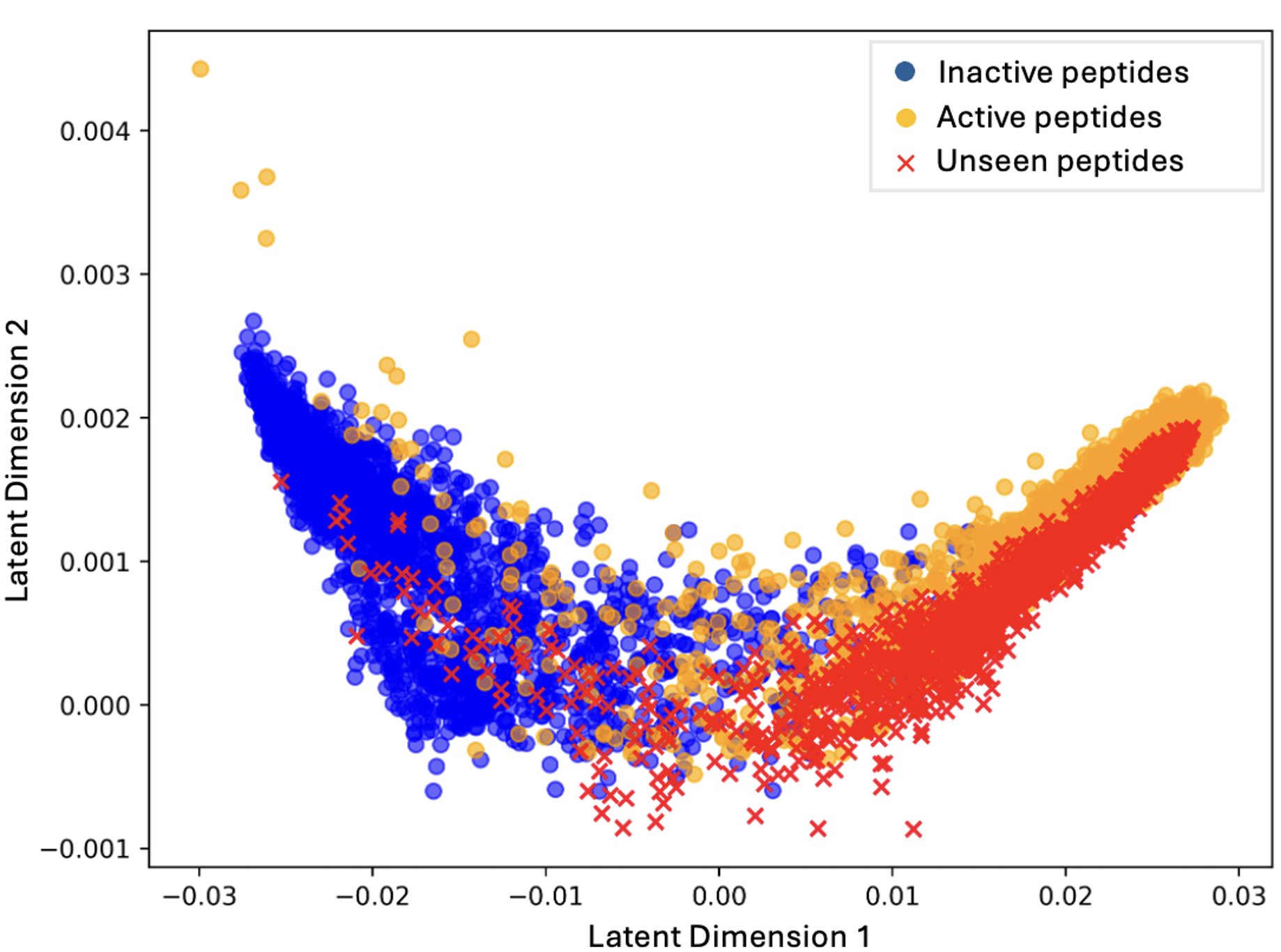}
    \caption{\FDD{} for AMP retrieval. Red points are held-out active peptide embeddings projected into the \FDD{} latent space.\label{fig:placeholder}}
\end{minipage}
\hfill
\begin{minipage}[t]{0.48\linewidth}
    \centering
    \includegraphics[width=\linewidth]{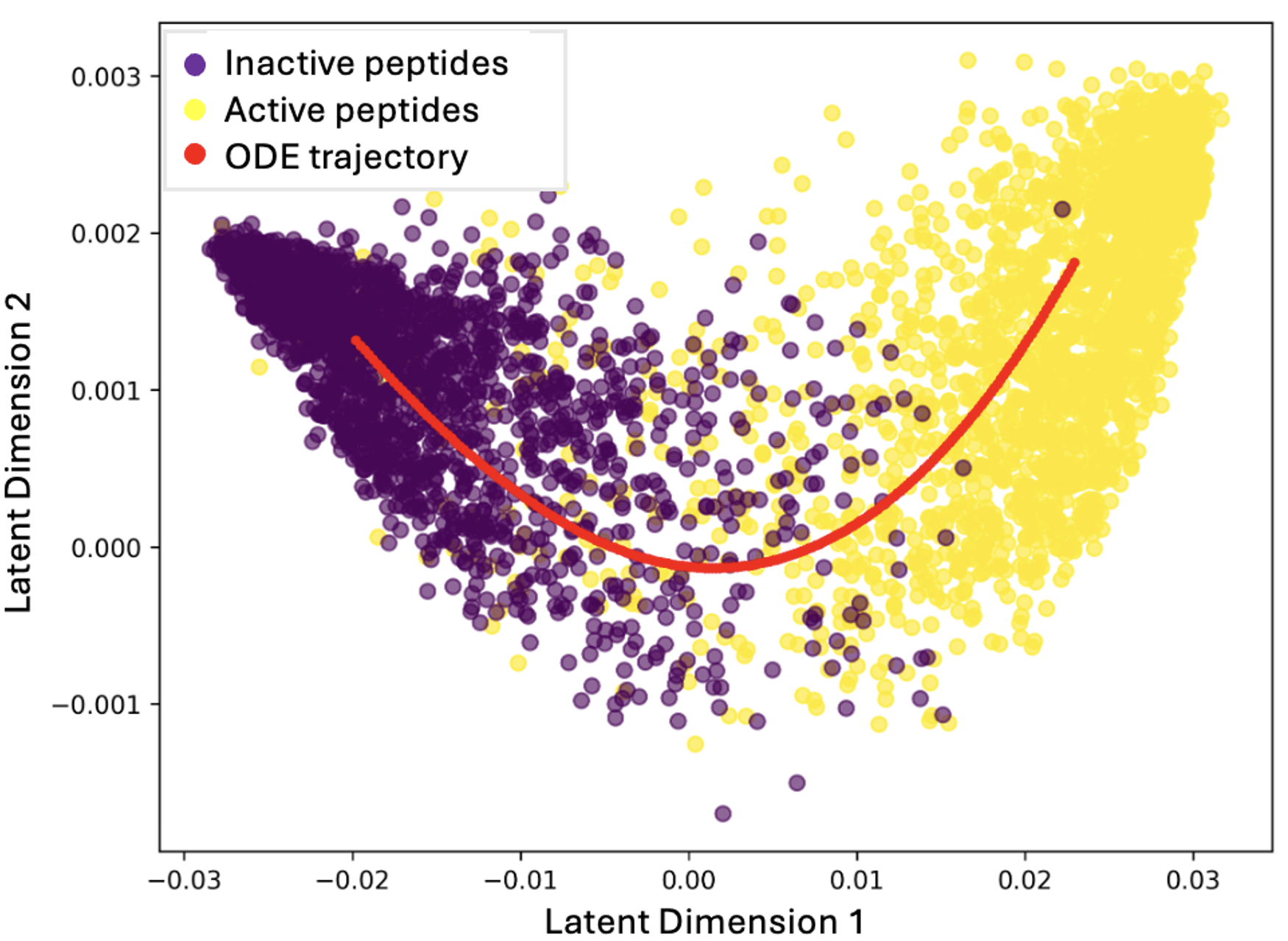}
    \caption{Interpolated trajectory in the \FDD{} latent space between non-AMP and AMP regions.\label{fig:noAMP/AMP}}
\end{minipage}
\vspace{-18pt}
\end{figure}

\FDD{} attains the highest $P_{\mathrm{amp}}$ ($0.90$; Table~\ref{tab:test_metrics}), on par with the strongest baselines while operating on representations not trained for classification. The embedding space separates active from inactive peptides (Figures~\ref{fig:placeholder} and~\ref{fig:umap_activity}), supporting accurate nearest-neighbor retrieval of novel actives.

\subsection{\FDD{} enables smooth latent-space interpolation}\label{section:experimental-results:interpolation}

Finally, we investigate whether \FDD{} enables smooth latent-space interpolation between AMP and non-AMP peptides. For this, we use the dataset from~\citet{Veltri2018-pl} and map all peptides into the \FDD{} latent representation space. Specifically, we select a non-AMP latent vector $\z_0$ and integrate a continuous trajectory from $\z_0$ towards the AMP manifold using a Neural ODE~\citep{chen2019neuralordinarydifferentialequations} trained with Sinkhorn loss~\citep{dimarino2020optimaltransportlossessinkhorn} to align the evolving distribution with that of active peptides (Figure~\ref{fig:noAMP/AMP}). Intermediate points along the trajectory are decoded back to peptide sequences via a learned MLP mapping, enabling analysis of physicochemical descriptor changes along the path.

\begin{figure}[htbp]
    \vspace{-5pt}
    \centering
    \includegraphics[width=1.0\linewidth]{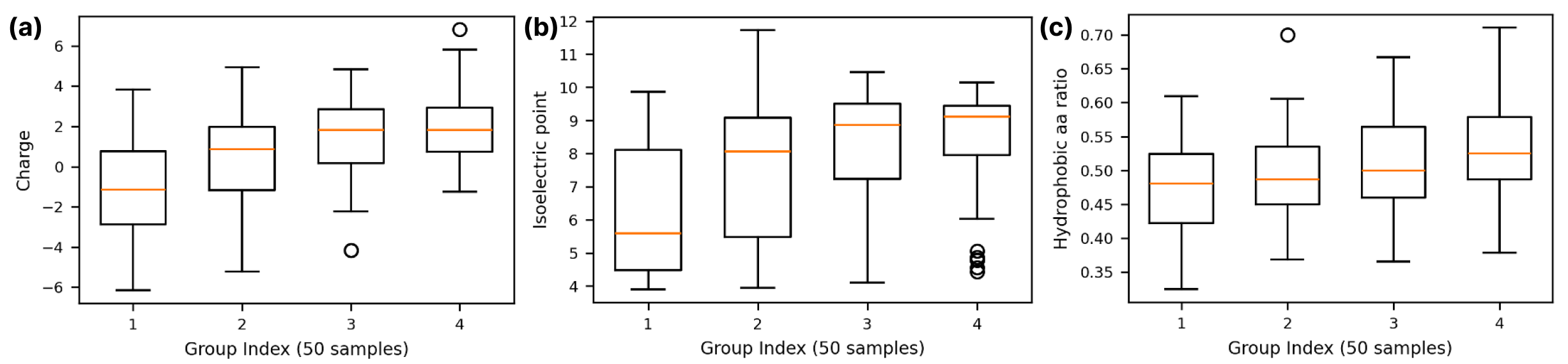}
    \caption{Evolution of physicochemical descriptors along the latent interpolation path:
\textbf{(a)} net charge, \textbf{(b)} isoelectric point, and \textbf{(c)} hydrophobic amino-acid ratio. All descriptors steadily increase across the interpolation path, indicating a smooth transition from non-AMP to AMP-like regions in latent space.}
    \label{fig:descriptors_full}
\vspace{-20pt}
\end{figure}

Peptides decoded along the learned interpolation path exhibit a smooth and monotonic transition in physicochemical descriptors that correlate with antimicrobial properties of peptides (Figure~\ref{fig:descriptors_full}). As an example, net charge increases from near-neutral to cationic (Figure~\ref{fig:descriptors_full} (a)), isoelectric point rises accordingly (Figure~\ref{fig:descriptors_full} (b)), and the hydrophobic amino-acid ratio grows, reflecting stronger membrane affinity typical of decoded AMPs (Figure~\ref{fig:descriptors_full} (c)). These results indicate that the \FDD{} latent space encodes a biologically meaningful manifold, where interpolation traces a gradual acquisition of antimicrobial properties.

\section{Related Work}
\label{sec:related}

We relate \FDD{} to prior work on adapting pretrained embeddings and on diffusion-geometric representation learning.

\paragraph{Adapting pretrained embeddings}
Two paradigms dominate transfer from pretrained embeddings. Fine-tuning updates the backbone and is expressive, but reshapes the pretrained geometry, degrades out-of-distribution robustness, and risks catastrophic forgetting~\citep{kumar2022, andreassen2021, rajaee2021}. Parameter-efficient variants such as adapters~\citep{houlsby2019adapters} and LoRA~\citep{hu2021lora} reduce cost but still modify the backbone. Probing instead freezes the backbone and fits a shallow predictor on top, preserving the geometry but lacking the expressivity for nonlinear, task-specific structure~\citep{belinkov2022, hewitt2019, white2021}. \FDD{} keeps the backbone frozen and adapts its embeddings through supervised diffusion, recovering fine-tuning expressivity.

\paragraph{Diffusion-geometric representation learning}
Diffusion Maps~\citep{coifman2006diffusion} build a diffusion operator from sample affinities and extract coordinates that preserve global structure while denoising local neighborhoods. PHATE~\citep{Moon2019-ue} extends this to a diffusion-potential representation stabilizing local and global structure. Supervised variants inject task information into the kernel. RF-GAP and RF-PHATE~\citep{rhodes2023rfgap, rhodes2023rfphate} derive label-aware affinities from a random forest, biasing the manifold toward task-relevant directions, and random forest autoencoders (RF-AE)~\citep{aumon2025rfae} instantiate the geometry-regularized autoencoder framework~\citep{duque2023grae} with a parametric encoder that extends the geometry to unseen samples.

\section{Conclusions}

We introduced \textsc{Freeze, Diffuse, Decode} (\FDD), a framework based on diffusion geometry that adapts pretrained transformer embeddings to downstream tasks without backbone fine-tuning. We showed that \FDD{} yields low-dimensional, predictive, and interpretable representations that improve over the pretrained embeddings on property prediction, retrieval and latent-space interpolation. As building the diffusion geometry scales cubically in the number of training points, large datasets require a landmark approximation, making \FDD{} best suited to the low-data regimes typical of antimicrobial peptide design.

\paragraph{Acknowledgments }
This project has received funding from the European Research
Council (ERC) under the European Funding Union’s Horizon 2020 research and innovation programme (grant agreement No 810115–DOG-AMP) [E.S.]; a FRQNT doctoral
scholarship and Mitacs-Globalink travel award [M.L.]; a Canada CIFAR AI chair (via
Mila) and Humboldt research fellowship [G.W.]; NSF grant 221232 [K.M.]; and an IVADO visiting scholar award
[K.M.].

\paragraph{Disclosure of Interests}  Projects in Szczurek lab at the University of Warsaw are
co-funded by Merck KGaA [E.S.].Part of this work was supported by the Helmholtz International Lab Causal Cell Dynamics (InterLabs-0029) - Grant support from the Initiative and Networking Fund of the Hermann von Helmholtz-Association Deutscher Forschungszentren e.V.[G.W.]

\newpage
\bibliography{main}
\newpage
\appendix

\section{ESKAPE Dataset Details}\label{data}

Antimicrobial Resistance (AMR) is among the most pressing threats to global health. As per \citep{Antimicrobial_Resistance_Collaborators2022-fa}, AMR was responsible for an estimated 1.27 million deaths and associated with a further 4.95 million, exceeding the toll of HIV/AIDS or malaria. AMR-related infections are estimated to kill around 10 million people annually by the year 2050~\citep{20173071720}. This burden is compounded by a stagnant discovery pipeline, with few mechanistically novel antibiotic classes reaching the clinic in decades while resistance to existing drugs continues to spread.

A disproportionate share of this burden traces to a small group of organisms. The acronym
\textbf{ESKAPE} 
\emph{Enterococcus faecium},
\emph{Staphylococcus aureus},
\emph{Klebsiella pneumoniae},
\emph{Acinetobacter baumannii},
\emph{Pseudomonas aeruginosa},
and \emph{Enterobacter} denotes pathogens that most readily
\emph{escape} the action of available antibiotics through multidrug resistance \citep{Rice2008-po}. These species dominate hospital-acquired infections and a
large fraction of attributable AMR mortality: in the 2019 estimates, six leading pathogens
(largely overlapping with this group) accounted for roughly 73\% of deaths directly
attributable to resistance \citep{Antimicrobial_Resistance_Collaborators2022-fa}, and they remain top priorities for
new antibacterials.
We evaluate antimicrobial activity on this panel, with one adjustment reflecting the
mortality ranking: we substitute \emph{Escherichia coli}, the single largest contributor
to AMR-attributable deaths~\citep{Antimicrobial_Resistance_Collaborators2022-fa}, for \emph{Enterobacter}~spp.,
yielding six clinically prioritized targets spanning Gram-positive (\emph{E.~faecium},
\emph{S.~aureus}) and Gram-negative (\emph{K.~pneumoniae}, \emph{A.~baumannii},
\emph{P.~aeruginosa}, \emph{E.~coli}) organisms. Antimicrobial peptides offer a promising
modality against these pathogens, but designing them requires predicting potency measured
as the minimum inhibitory concentration (MIC) from sequence, and labeled MIC data are
scarce on a per-species basis, precisely the low-data regime our method targets.
Each peptide is labeled as \emph{active} if its measured MIC is below $32\,\mu$M.
Sequences are split 80/20 (train/test) using stratified sampling (fixed seed),
with no duplicate sequences present in the data.
The full dataset comprises \textbf{14{,}843} unique peptides
(11{,}872 train / 2{,}971 test across all species).

\begin{table}[h]
\centering
\caption{Per-species statistics for the ESKAPE MIC dataset.
$N$ is the number of unique peptides. Peptides with MIC $<32\,\mu$M are labeled active.}
\label{tab:eskape_dataset}
\scriptsize
\setlength{\tabcolsep}{3pt}
\renewcommand{\arraystretch}{0.9}
\begin{tabular}{lrrrrr}
\toprule
\textbf{Species} & $N$ & $N_{\mathrm{tr}}$ & $N_{\mathrm{te}}$ &
\textbf{Active tr/te} & \textbf{\% Act.} \\
\midrule
\textit{E. faecium}     &   293 &   234 &  59 & 161 / 40  & 69\% \\
\textit{S. aureus}      & 4{,}264 & 3{,}411 & 853 & 1{,}985 / 497 & 58\% \\
\textit{K. pneumoniae}  & 1{,}384 & 1{,}107 & 277 & 611 / 153 & 55\% \\
\textit{A. baumannii}   & 1{,}177 &   941 & 236 & 656 / 165 & 70\% \\
\textit{P. aeruginosa}  & 3{,}101 & 2{,}480 & 621 & 1{,}393 / 349 & 56\% \\
\textit{E. coli}        & 4{,}624 & 3{,}699 & 925 & 2{,}440 / 610 & 66\% \\
\midrule
\textbf{Total}          & \textbf{14{,}843} & \textbf{11{,}872} & \textbf{2{,}971} & -- & -- \\
\bottomrule
\end{tabular}
\end{table}

\section{Local Intrinsic Dimensionality}\label{apd:id}

A representation that resolves a task with fewer effective dimensions exposes a simpler, smoother manifold along which to interpolate, retrieve neighbors, and fit a head from few labels. ID quantifies the geometric cost of adaptation: fine-tuning recruits additional directions to realign the backbone, and in the low-data AMP regime this added complexity is what drives overfitting.

For each species we restrict to \emph{active} peptides ($\mathrm{MIC}<32\,\mu$M, $y{=}1$) after removing duplicate sequences, so that ID characterizes the geometry of the functionally relevant region of the embedding space. We estimate ID locally via the participation ratio (PR;~\citet{Gao214262}) of the local PCA spectrum, a continuous effective-rank measure. For each active point $x_i$ we take its $k{=}20$ nearest neighbors under exact ($\ell_2$, brute-force) distance within the active subset, center them by their mean to form $N_i\in\mathbb{R}^{k\times d}$, and compute the local covariance
\begin{equation*}
  C_i = \frac{1}{k-1}\,N_i^\top N_i .
\end{equation*}
The local effective dimensionality is
\begin{equation*}
  \mathrm{PR}_i = \frac{\big(\sum_j \lambda_j\big)^2}{\sum_j \lambda_j^2},
\end{equation*}
which equals $1$ for a strictly one-dimensional structure and $d$ for a perfectly isotropic $d$-dimensional cloud, where $\lambda_j$ are the eigenvalues of $C_i$. The ID of an embedding space is the mean over active points, $\widehat{\mathrm{ID}}=\tfrac{1}{n_{\mathrm{act}}}\sum_{i=1}^{n_{\mathrm{act}}}\mathrm{PR}_i$.

\paragraph{Embedding spaces}
ID is computed in three spaces per species:
\begin{itemize}\setlength{\itemsep}{1pt}
  \item \textbf{PT}: mean-pooled frozen pretrained Hyformer embeddings;
  \item \textbf{\textsc{FDD}}: the \textsc{FDD} projection of the PT embeddings;
  \item \textbf{FT}: mean-pooled embeddings from the fine-tuned Hyformer.
\end{itemize}

\paragraph{Validity of the local scale}
PR estimates \emph{local, linear} dimensionality, the effective rank of the tangent-space covariance, and is bounded above by $k-1=19$. The reported values ($\sim$5--8) sit well inside this range, so $k$ is large enough to resolve the relevant scale while remaining local enough to approximate the tangent space.

\section{Experimental Details}\label{apd:method-details}

This appendix collects the hyperparameters and implementation choices deferred from~\Cref{sec:methods}.

\paragraph{Random forest and proximities}
We fit a random forest with $100$ trees, unlimited maximum depth, and a minimum of $1$ sample per leaf to the frozen training embeddings $\{(\h_n,y_n)\}$ without feature standardization (random forests are invariant to monotone input transformations). RF-GAP proximities~(\Cref{eq:rfgap}) are symmetrized as $\tfrac12(\mathbf{W}+\mathbf{W}^\top)$ and row-normalized into the diffusion operator $\mathbf{P}$.

\paragraph{RF-PHATE targets}
We raise $\mathbf{P}$ to diffusion time $t = \texttt{auto}$, selected by the von Neumann entropy criterion of PHATE~\citep{Moon2019-ue}, apply the potential transform, and embed the potential distances into $\R^d$ with $d \in \{16,32,64\}$ (selected by cross-validated AUC; $d=32$ fixed for all visualizations) using landmark metric MDS with an SGD solver. Up to $L = \min(n_{\mathrm{train}}-1,\,2000)$ landmarks are used to make the diffusion graph tractable; the diffusion operator includes a teleportation term with $\beta = 0.9$.

\paragraph{Geometry-regularized autoencoder}
The encoder and decoder are $3$-layer multilayer perceptrons of widths $[800, 400, 100]$ with ELU activations; the decoder ends in a log-softmax over the $L$ landmark proximities. We sweep the loss weight $\lambda \in \{10^{-3}, 10^{-2}, 10^{-1}\}$ in~\Cref{eq:rfae-loss} (cross-validated; $\lambda = 10^{-2}$ is typically best) and optimize with AdamW (learning rate $10^{-3}$, weight decay $10^{-5}$, batch size $64$, $500$ epochs).

\paragraph{Latent-space interpolation}
The neural-ODE velocity field $v_\psi$ is a \texttt{2}-layer network trained with the Sinkhorn divergence at regularization $\varepsilon = 0.1$, integrated with \texttt{rk4} and the adjoint method. We give the full entropic optimal-transport objective in~\Cref{sec:decode}.

\subsection{Embedding Extraction Details}\label{emb:ext}
All the embeddings are extracted from the frozen backbone. For sequences of length $T$, we pool the final-layer per-token hidden states
$\mathbf{H} = [\mathbf{h}_1, \dots, \mathbf{h}_T]$ using a binary attention mask
$m_t \in \{0,1\}$ that excludes padding. We use two pooling operators:
\begin{align*}
  \text{CLS:}  \quad & \mathbf{h}_0, \\
  \text{Mean:} \quad & \bar{\mathbf{h}} = \frac{\sum_t m_t\, \mathbf{h}_t}{\sum_t m_t}.
\end{align*}

\paragraph{Hyformer (HYF) Embeddings}

We load a pretrained Hyformer: an 8-layer transformer with
$d{=}512$, 8 attention heads, and an amino-acid vocabulary of 34 tokens. Sequences are
tokenized with the ESM amino-acid tokenizer (padded or truncated to 50 tokens) and
passed through the backbone to obtain hidden states
$\mathbf{H} \in \mathbb{R}^{T \times 512}$. We derive two representations: \textbf{PT-CLS},
the hidden state $\mathbf{h}_0 \in \mathbb{R}^{512}$ at position~0 (the task/CLS token),
and \textbf{PT-Mean}, the mask-weighted mean $\bar{\mathbf{h}} \in \mathbb{R}^{512}$.

\paragraph{ESM2 Embeddings}

We load ESM2-35M (12 layers, $d{=}480$) from
Hugging~Face. Sequences are tokenized with the ESM2 tokenizer (padded or truncated to
128 tokens, sufficient for antimicrobial peptides) and passed through the backbone. To
mirror the Hyformer PT-Mean protocol, we use mean pooling only, yielding
$\bar{\mathbf{h}} \in \mathbb{R}^{480}$ over the final-layer hidden states.

\begin{table}[h]
\centering
\caption{Frozen-embedding extraction configuration.}
\label{tab:embed_config}
\small
\begin{tabular}{lcccccc}
\toprule
\textbf{Backbone} & \textbf{Layers} & $d$ & \textbf{Max tokens} & \textbf{Pooling} & \textbf{Batch} \\
\midrule
Hyformer-33M  &  8 &  512 &  50 & CLS, Mean & 64 \\
ESM2-35M & 12 & 480 & 128 & CLS, Mean      & 16 \\
\bottomrule
\end{tabular}
\end{table}

\paragraph{Downstream classifiers}

In both cases, the frozen embeddings serve as input to a logistic regression
classifier with $\ell_2$ regularization, fit on the 80\,\% training split.
The regularization strength $C$ is selected via 5-fold cross-validation over
a log-uniform grid $[10^{-4}, 10^{4}]$.
Features are standardized (zero mean, unit variance) before fitting,
with the scaler fitted on training data only.
For the fine-tuning conditions (\textbf{FT}), the full model backbone
(Hyformer or ESM2) plus a linear head is end-to-end trained using AdamW
($\text{lr}{=}10^{-4}$, $\beta{=}(0.9,0.95)$, cosine schedule with 10\,\%
linear warm-up, early stopping on validation AUC with patience equal to 15).

\section{Ablations}\label{sec:ablations}

\paragraph{FDD latent dimension sensitivity}
All reported results use a fixed latent dimension $d{=}32$ and regularization
$\lambda{=}10^{-2}$ (Table~\ref{tab:hyformer_esm2_auc}); this section varies each in turn to
assess sensitivity. We sweep $d \in \{2,8,16,32,64,128\}$ at fixed $\lambda{=}10^{-2}$ on
frozen Hyformer mean embeddings ($d_{\text{in}}{=}512$); results are mean (std) over 5
seeds (Table~\ref{tab:rfae_dim_sweep}).

Performance is flat in $d$ above a small threshold: for the five larger species it saturates by
$d{=}16$, with the fixed default $d{=}32$ within one standard deviation of the per-species best
in every case (e.g.\ \textit{A. baumannii} and \textit{P. aeruginosa} peak exactly at $d{=}32$;
\textit{S. aureus}, \textit{E. coli}, and \textit{K. pneumoniae} differ by ${\leq}0.003$ AUC).
The sole exception is \textit{E. faecium}, our smallest dataset ($N_{\text{tr}}{=}234$), which
peaks at $d{=}8$ ($0.853$ vs.\ $0.834$ at $d{=}32$); the optimal latent dimension shrinks with
the available data, consistent with the low-data regime \FDD{} targets. Performance drops sharply
only at $d{=}2$, indicating \FDD{} needs a handful of dimensions but no more. We therefore fix
$d{=}32$ as a robust default across species.
\begin{table}[h]
\centering
\caption{\FDD{} dimensionality sweep on Hyformer embeddings across ESKAPE bacterial
  species (ROC-AUC). The best $d$ per species is marked \textbf{bold}. Mean (std) across
  5 seeds; $\lambda{=}10^{-2}$ fixed throughout.}
\label{tab:rfae_dim_sweep}
\resizebox{\textwidth}{!}{%
\begin{tabular}{l rrrrrr}
\toprule
\textbf{Species}
  & $d{=}2$ & $d{=}8$ & $d{=}16$ & $d{=}32$ & $d{=}64$ & $d{=}128$ \\
\midrule
\textit{A.~baumannii}
  & 0.696 (0.016) & 0.846 (0.015) & 0.870 (0.011)
  & \textbf{0.871 (0.013)} & 0.869 (0.017) & 0.866 (0.013) \\
\textit{E.~faecium}
  & 0.848 (0.029) & \textbf{0.853 (0.013)} & 0.843 (0.012)
  & 0.834 (0.014) & 0.833 (0.017) & 0.836 (0.018) \\
\textit{K.~pneumoniae}
  & 0.785 (0.002) & 0.821 (0.009) & 0.829 (0.007)
  & 0.828 (0.010) & 0.828 (0.010) & \textbf{0.830 (0.011)} \\
\textit{P.~aeruginosa}
  & 0.689 (0.006) & 0.843 (0.004) & 0.852 (0.003)
  & \textbf{0.859 (0.004)} & 0.857 (0.004) & 0.853 (0.006) \\
\textit{S.~aureus}
  & 0.717 (0.003) & 0.817 (0.003) & \textbf{0.835 (0.006)}
  & 0.832 (0.009) & 0.832 (0.008) & 0.830 (0.006) \\
\textit{E.~coli}
  & 0.763 (0.010) & 0.858 (0.008) & \textbf{0.868 (0.005)}
  & 0.867 (0.004) & 0.867 (0.008) & 0.866 (0.005) \\
\bottomrule
\end{tabular}}
\end{table}

\paragraph{FDD geometric regularization sensitivity}
Holding latent dimensionality $d{=}32$ fixed, we sweep the geometric regularization hyperparameter
$\lambda \in \{0.0,10^{-3},10^{-2},10^{-1},1\}$
(Table~\ref{tab:rfae_lam_sweep}; PT is the frozen 512-d Hyformer baseline). \FDD{} improves over
the frozen baseline on five of six species, the exception being \textit{E. coli}
(PT $0.882$ vs.\ \FDD{} $0.867$), where the raw embeddings are already highly discriminative
(consistent with Table~\ref{tab:hyformer_esm2_auc}). Sensitivity to $\lambda$ is uniformly low:
the full grid moves AUC by ${\leq}0.015$ for every species, and typically by ${\leq}0.005$, so
any $\lambda \in [10^{-3},10^{-1}]$ performs within noise. We fix $\lambda{=}10^{-2}$ throughout.

\begin{table}[h!]
\centering
\caption{\FDD{} ROC-AUC across five regularization strengths $\lambda$ at fixed latent dimensionality $d{=}32$, evaluated within each ESKAPE bacterial species. PT is the frozen Hyformer mean-embedding baseline. The best $\lambda$ per species is
  marked \textbf{bold}. Mean (std) across 5 seeds.}
\label{tab:rfae_lam_sweep}
\resizebox{\linewidth}{!}{%
\begin{tabular}{l cc ccccc l}
\toprule
\textbf{Species} & \textbf{PT} &
  $\lambda{=}0$ & $\lambda{=}10^{-3}$ & $\lambda{=}10^{-2}$ & $\lambda{=}10^{-1}$ & $\lambda{=}1$ &
  \textbf{Best $\lambda$} \\
\midrule
\textit{A.~baumannii}  & 0.843 &
  0.841 (0.010) & 0.856 (0.011) & \textbf{0.871 (0.013)} & 0.869 (0.014) & 0.863 (0.009) & $10^{-2}$ \\
\textit{E.~faecium}    & 0.800 &
  0.799 (0.035) & \textbf{0.842 (0.021)} & 0.834 (0.014) & 0.828 (0.019) & 0.833 (0.023) & $10^{-3}$ \\
\textit{K.~pneumoniae} & 0.826 &
  0.806 (0.009) & 0.824 (0.012) & 0.828 (0.010) & \textbf{0.830 (0.009)} & 0.828 (0.013) & $10^{-1}$ \\
\textit{P.~aeruginosa} & 0.842 &
  0.818 (0.004) & 0.854 (0.008) & \textbf{0.859 (0.004)} & 0.856 (0.003) & 0.853 (0.003) & $10^{-2}$ \\
\textit{S.~aureus}     & 0.823 &
  0.816 (0.002) & 0.831 (0.010) & 0.832 (0.009) & \textbf{0.835 (0.007)} & 0.835 (0.009) & $10^{-1}$ \\
\textit{E.~coli}       & 0.882 &
  0.864 (0.004) & 0.866 (0.009) & 0.867 (0.004) & \textbf{0.870 (0.005)} & 0.868 (0.002) & $10^{-1}$ \\
\bottomrule
\end{tabular}}
\end{table}

\begin{figure}[p]
\centering
\includegraphics[
  width=0.95\linewidth,
  height=0.8\textheight,
  keepaspectratio
]{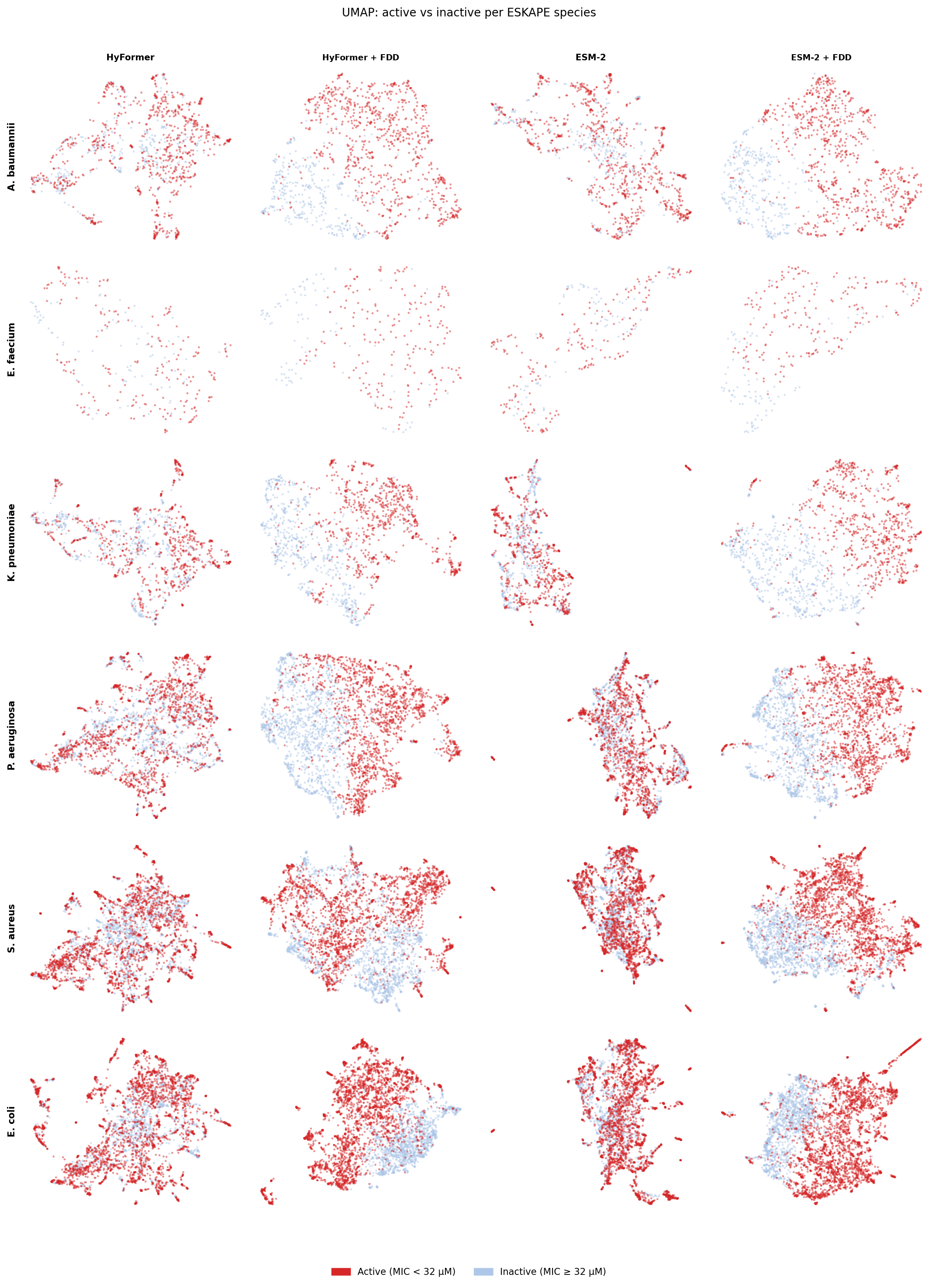}
\caption{UMAP of CLS embeddings per ESKAPE bacterial species, colored by antimicrobial activity
(red: active, MIC $<32\,\mu$M; blue: inactive). For each species we show frozen pretrained
embeddings and their \FDD{} projection ($32$-d), for both backbones. Each panel is an
independent UMAP fit. Only the within-panel arrangement of active vs.\ inactive points is
meaningful, not positions across panels.}
\label{fig:umap_activity}
\end{figure}

\section{Additional Results}\label{apd:first}

\subsection{\textsc{FDD} representations remain competitive on out-of-distribution molecular property prediction}

To evaluate whether our framework generalizes beyond antimicrobial peptides, we benchmark \FDD{} on out-of-distribution molecular property prediction tasks from~\citep{steshin2023hi}. We probe frozen embeddings of Hyformer with a $k$-nearest neighbors ($k\text{NN}$) classifier. Following~\cite{steshin2023hi}, we compare \textsc{FDD} against molecular fingerprints (ECFP4, MACCS), frozen Hyformer embeddings, and a label-frequency dummy baseline. Full end-to-end fine-tuning serves as an empirical upper bound.

\begin{table}[!h]
\caption{$k$NN classification performance on molecular property prediction.
Among $k$NN methods, the best result per dataset is marked \textbf{bold}.
Fine-tuning is reported separately as an upper-bound reference. Mean (std)
across three seeds.\label{tab:mol_prop}}
\vspace{-2ex}
\begin{center}
\begin{scriptsize}
\begin{sc}
\begin{tabular}{@{}lcccc@{}}
\toprule
& \multicolumn{4}{c}{Dataset, AUPRC ($\uparrow$)} \\
\cmidrule(lr){2-5}
Method & DRD2-Hi & HIV-Hi & KDR-Hi & Sol-Hi \\
\midrule
Dummy baseline       & 0.677 (0.061) & 0.040 (0.014) & 0.609 (0.081) & 0.215 (0.008) \\
ECFP4                & 0.706 (0.047) & 0.067 (0.029) & 0.646 (0.048) & 0.426 (0.022) \\
MACCS                & 0.702 (0.042) & \textbf{0.072 (0.036)} & 0.610 (0.072) & 0.422 (0.009) \\
Hyformer             & 0.700 (0.057) & 0.052 (0.018) & 0.646 (0.071) & 0.484 (0.040) \\
\FDD{} (ours)        & \textbf{0.747 (0.040)} & 0.049 (0.016) & \textbf{0.663 (0.058)} & \textbf{0.494 (0.054)} \\
\cmidrule(lr){2-5}
Fine-tuning          & 0.784 (0.082) & 0.158 (0.128) & 0.701 (0.022) & 0.640 (0.036) \\
\bottomrule
\end{tabular}
\end{sc}
\end{scriptsize}
\end{center}
\vspace{-2ex}
\end{table}

The empirical results demonstrate that \textsc{FDD} yields robust, competitive molecular representations, establishing the top frozen-feature performance on three of the four tasks. Broadly, these findings confirm that \textsc{FDD} remains competitive with established baselines outside the primary peptide domain.

\subsection{Sample Efficiency Under Label Scarcity}
\label{sec:sample-efficiency}
 
\begin{figure}[htbp]
  \centering
  \includegraphics[width=\textwidth]{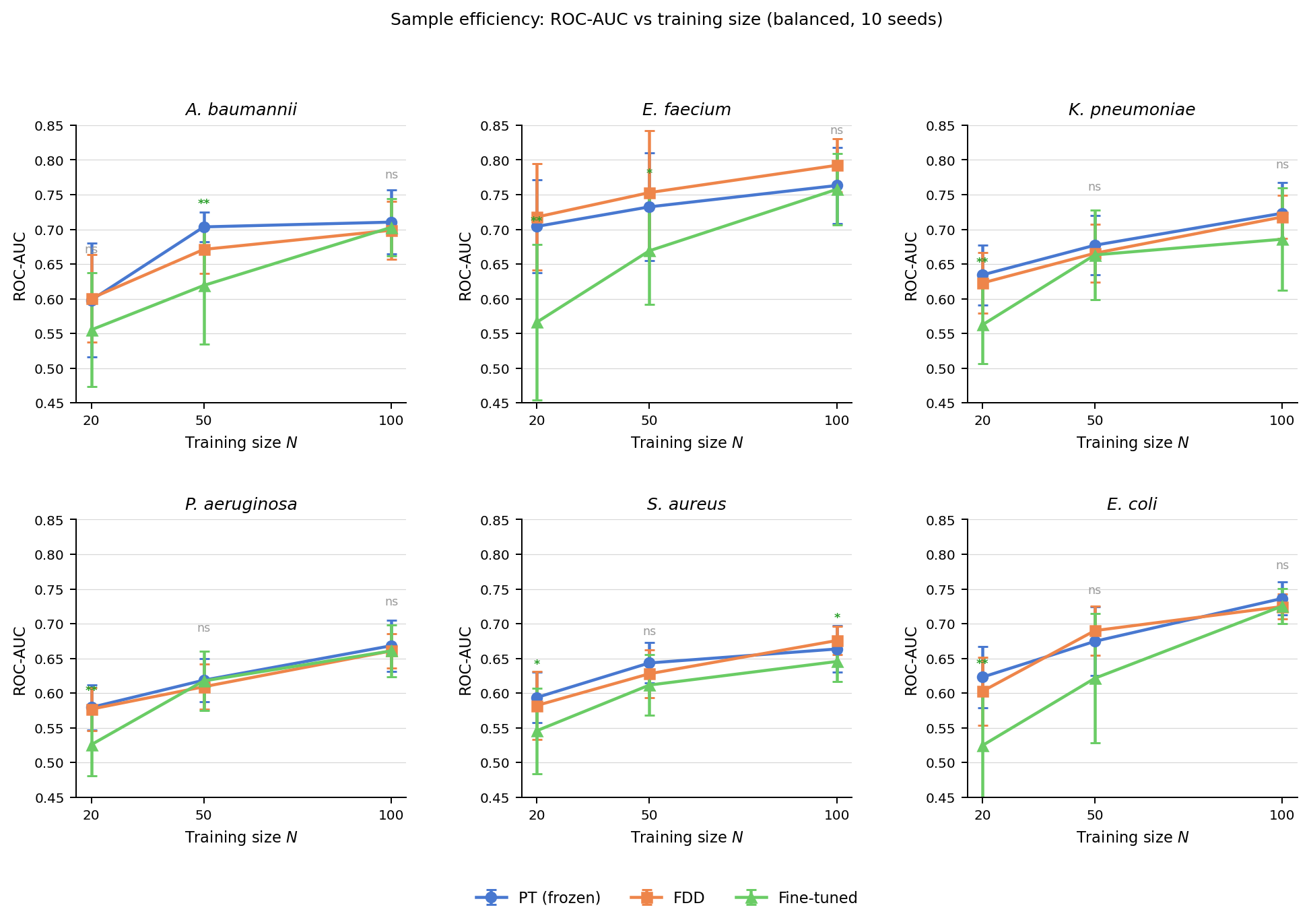}
  \caption{Sample efficiency across ESKAPE bacterial species. Mean ROC-AUC versus
  balanced training size $N$ over $10$ seeds (error bars $\pm 1$ s.d., shared
  axes). Markers report a Wilcoxon test of FT vs.\ PT
  ($^{*}\,p{<}0.05$, $^{**}\,p{<}0.01$, $^{***}\,p{<}0.001$, \emph{ns}~$=$~not significant).}
  \label{fig:sample-efficiency}
\end{figure}

Fine-tuning attains the best full-data AUC (Table~\ref{tab:hyformer_esm2_auc}), which invites
the question of why one would adapt frozen embeddings rather than fine-tune. The
answer is that fine-tuning is unreliable when labels are scarce. For each ESKAPE
species we subsample the training split to $N \in \{20, 50, 100\}$ with balanced
classes (fixed test split, $10$ seeds) and compare frozen pretrained embeddings
(PT), FDD, and end-to-end fine-tuning (FT) under an identical linear probe; we
test the FT-versus-PT gap with a one-sided Wilcoxon signed-rank test over paired
seeds (Figure~\ref{fig:sample-efficiency}).
 
In every species FT is the weakest method at small $N$ and converges to the
frozen methods only by $N=100$, trailing by up to ${\approx}\,0.14$ AUC at
$N=20$ (\emph{E. faecium}); the Wilcoxon test confirms FT $<$ PT at low-$N$
points, though fine-tuning's large variance at $N=20$ leaves several gaps short
of significance. FDD tracks PT throughout, and in five of six species the two
are statistically indistinguishable. The exception is \emph{E. faecium}, our
smallest set, where FDD separates from both, and its margin over PT widens with
$N$, matching Table~\ref{tab:hyformer_esm2_auc}.
 
This isolates one point: the method that is strongest at full data is the least
reliable under scarcity, motivating frozen adaptation. It is \emph{not} evidence
that FDD beats frozen embeddings at small $N$, where they are comparable; FDD's
edge is a full-data, geometric effect (Tables~\ref{tab:hyformer_esm2_auc}, \ref{tab:ID}), achieved while matching the frozen baseline's sample
efficiency.

\end{document}